\documentclass[sigconf]{acmart}
\AtBeginDocument{%
  }

\usepackage{algorithm}
\usepackage{algorithmic}
\usepackage{graphicx}
\usepackage{subfigure}
\usepackage{epstopdf}
\usepackage{xspace}
\usepackage{wrapfig}
\usepackage{amsthm}
\usepackage{newfloat}
\usepackage{listings}
\usepackage{xspace}
\usepackage{bibentry}
\usepackage{multirow}
\usepackage{subcaption} 
\usepackage{booktabs}
\usepackage{wrapfig}
\usepackage{tikz,xcolor}
\usepackage{scalerel}
\usepackage{hyperref}
\usepackage{pifont}
\usepackage{xcolor}
\usepackage{listings}
\usepackage{balance}
\lstdefinestyle{codeStyle}{
  language=Python,
  basicstyle=\ttfamily\small,
  numbers=left,
  numberstyle=\tiny\color{gray},
  backgroundcolor=\color{gray!10},
  keywordstyle=\color{blue},
  commentstyle=\color{green!50!black},
  stringstyle=\color{red!60!black},
  frame=single,
  breaklines=true,
  tabsize=2
}
\newcommand{\cmark}{\ding{51}} 
\newcommand{\xmark}{\ding{55}}

\copyrightyear{2026}
\acmYear{2026}
\setcopyright{cc}
\setcctype{by}
\acmConference[KDD '26]{Proceedings of the 32nd ACM SIGKDD Conference on Knowledge Discovery and Data Mining V.1}{August 09--13, 2026}{Jeju Island, Republic of Korea}
\acmBooktitle{Proceedings of the 32nd ACM SIGKDD Conference on Knowledge Discovery and Data Mining V.1 (KDD '26), August 09--13, 2026, Jeju Island, Republic of Korea}
\acmPrice{}
\acmDOI{10.1145/3770854.3785681}
\acmISBN{979-8-4007-2258-5/2026/08}

\newcommand{\benchname}{HAROOD\xspace}
\begin{document}

\title{HAROOD: A Benchmark for Out-of-distribution Generalization in Sensor-based Human Activity Recognition}

\author{Wang Lu}
\orcid{0000-0003-4035-0737}
\affiliation{%
  \institution{William \& Mary}
  \department{Department of Data Science}
  \city{Williamsburg}
  \state{Virginia}
  \country{United States}
}
\email{newlw230630@gmail.com}

\author{Yao Zhu}
\orcid{0000-0003-0991-1970}
\affiliation{%
  \institution{William \& Mary}
  \department{Department of Data Science}
  \city{Williamsburg}
  \state{Virginia}
  \country{United States}
}
\email{eezhuy@gmail.com}

\author{Jindong Wang}
\orcid{0000-0002-4833-0880}
\authornote{Correspondence to: Jindong Wang. }
\affiliation{%
  \institution{William \& Mary}
  \department{Department of Data Science}
  \city{Williamsburg}
  \state{Virginia}
  \country{United States}
}
\email{jdw@wm.edu}


\begin{abstract}
Sensor-based human activity recognition (HAR) mines activity patterns from the time-series sensory data.
In realistic scenarios, variations across individuals, devices, environments, and time introduce significant distributional shifts for the same activities.
Recent efforts attempt to solve this challenge by applying or adapting existing out-of-distribution (OOD) algorithms, but only in certain distribution shift scenarios (e.g., cross-device or cross-position), lacking comprehensive insights on the effectiveness of these algorithms.
For instance, is OOD necessary to HAR? Which OOD algorithm performs the best? 
In this paper, we fill this gap by proposing \benchname, a comprehensive benchmark for HAR in OOD settings. We define 4 OOD scenarios: cross-person, cross-position, cross-dataset, and cross-time, and build a testbed covering 6 datasets, 16 comparative methods (implemented with CNN-based and Transformer-based architectures), and two model selection protocols. Then, we conduct extensive experiments and present several findings for future research, e.g., no single method consistently outperforms others, highlighting substantial opportunity for advancement.
Our codebase is highly modular and easy to extend for new datasets, algorithms, comparisons, and analysis, with the hope to facilitate the research in OOD-based HAR. 
Our implementation is released and can be found at \url{https://github.com/AIFrontierLab/HAROOD}.
\end{abstract}

\ccsdesc[500]{Computing methodologies~Supervised learning by classification}

\keywords{Human Activity Recognition; Transfer Learning; Domain Generalization; Out-of-Distribution; Benchmark}


\maketitle

\section{Introduction}

\begin{figure*}[!thbp]
  \centering
  \subfigure[Challenge in reality.]{
    \includegraphics[height=0.2\textwidth]{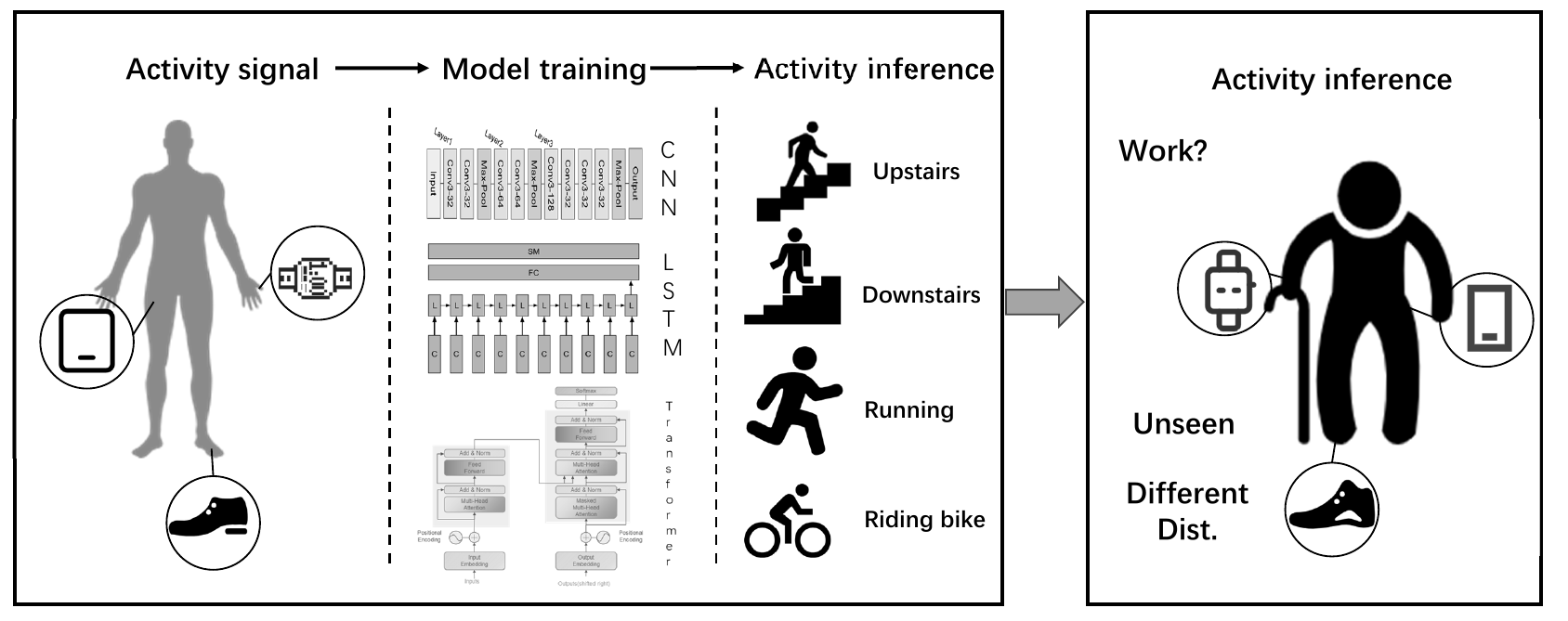}
    \label{fig:intro-problem}
  }
  \subfigure[Four scenarios.]{
    \includegraphics[height=0.2\textwidth]{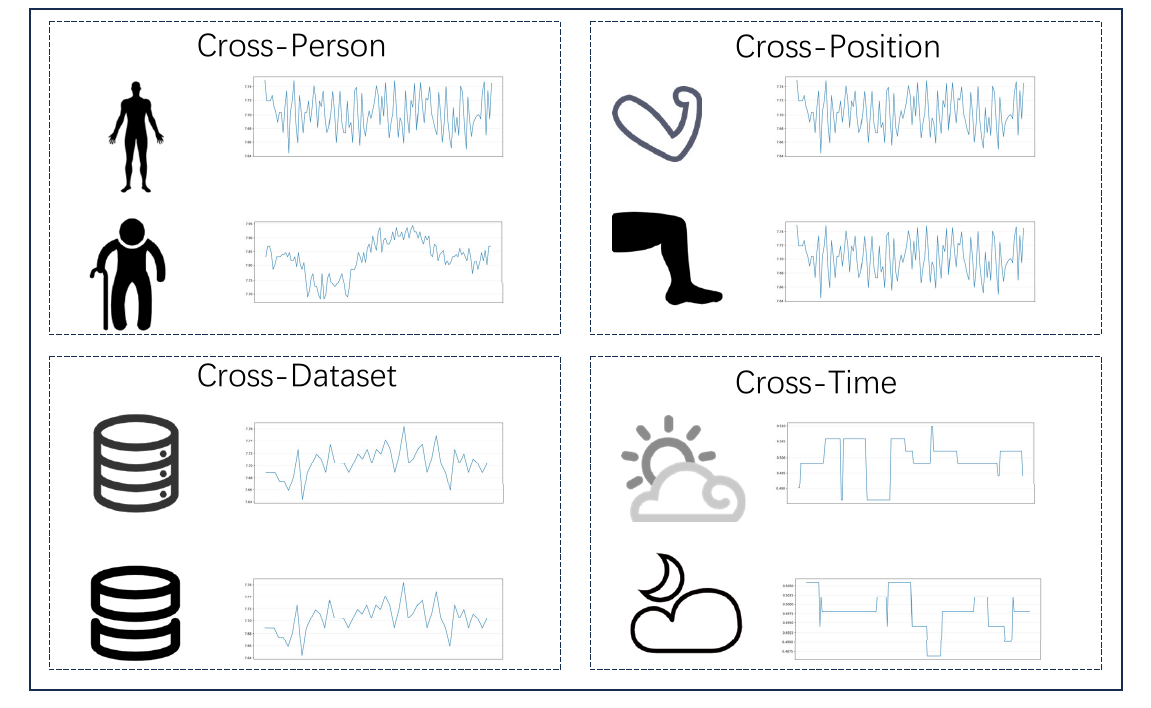}
    \label{fig:frame-scenari}
  }
  \vspace{-0.6cm}
  \caption{This figure illustrates our problem and the four evaluation scenarios. (a) Collecting data from young individuals and training models to infer their behavior does not guarantee that the same models will perform well when applied to older adults, whose behaviors follow different distributions and are unseen during training. (b)Illustrative diagrams and corresponding data samples for the four scenarios.}
  \label{fig:intro}
  \vspace{-.6cm}
\end{figure*}

Machine learning systems for sensor-based human activity recognition (HAR) play a central role in many critical real-world applications such as healthcare monitoring~\cite{jubair2025machine}, assisted living~\cite{teng2023pdges}, fitness tracking~\cite{liu2024imove}, smart homes~\cite{thukral2025layout}, and interactive environments~\cite{leng2024imugpt}. 
These systems rely on data from wearable or ambient sensors—such as accelerometers, gyroscopes, IMUs—to infer user behavior in a continuous, privacy-preserving, and non-invasive manner~\cite{fawaz2019deep,wang2019deep}.

However, in real-world sensor-based applications, inference performance often deteriorates sharply when models are deployed in unseen environments, just as shown in \figureautorefname~\ref{fig:intro-problem}. 
Factors such as user-specific biomechanics, device hardware variations, sensor placement (e.g., wrist vs. pocket), environmental differences (indoor vs. outdoor), and temporal changes (e.g., sensor drift) introduce distribution shifts between training and test domains. 
For example, it is neither feasible nor ethical to induce a large number of elderly individuals to fall for the purpose of training a elderly fall-detection model.
A model trained on one context may thus generalize poorly to another unseen setting, a phenomenon known in the literature as out‑of‑distribution (OOD) generalization~\cite{lu2022semantic}.
Academically, this challenge is formalized under the umbrella of OOD/Domain Generalization (DG), which aims to learn representations that are invariant across multiple source domains, such that the resulting model performs well on new, unseen target domains without any access to target-domain data during training~\cite{zhou2022domain,wang2022generalizing}.

According to \cite{wang2022generalizing}, existing OOD methods fall into three broad categories: data manipulation, representation learning, and learning strategy. Data manipulation methods—such as Mixup~\cite{zhang2018mixup}, domain-specific augmentations~\cite{li2021progressive}, and style randomization~\cite{huang2017arbitrary}—increase training diversity by synthesizing novel samples across source domains. Representation learning approaches align feature distributions via adversarial learning~\cite{ganin2016domain}, maximum mean discrepancy (MMD)~\cite{li2018domain}, CORAL~\cite{sun2016deep}, or instance‑level normalization~\cite{pan2018two}, promoting domain-invariant representations. Finally, learning strategy techniques employ meta-learning setups~\cite{li2018learning}, distributionally robust optimization~\cite{sagawadistributionally}, or ensemble/self-supervised schemes~\cite{rame2022diverse} to build models resilient to unseen target domains.

Several works attempted to apply OOD methods to HAR. For instance, AFFAR~\cite{qin2022domain} fused domain-invariant and domain-specific representations in a unified model, achieving significant generalization improvements across multiple public HAR datasets and even in an ADHD diagnosis scenario—without accessing any target-domain data during training. 
A recent work, CCIL (Categorical Concept Invariant Learning), proposed a concept‑matrix-based regularization that enforces both feature-level and logit-level invariance for samples within the same activity class, significantly enhancing cross-person, cross-dataset, and cross-position generalization in sensor‑based human activity recognition~\cite{xiong2025generalizable}.

Although methods like LAG~\cite{lu2022local} have been evaluated on two public HAR datasets (e.g., USC-HAD~\cite{zhang2012usc} and DSADS~\cite{barshan2014recognizing}) and show an impressive average accuracy improvement under cross-person evaluation settings, some efforts have aggregated datasets into more structured benchmarks. 
For example, the DAGHAR benchmark~\cite{napoli2024benchmark} standardized six smartphone-based inertial datasets—including UCI‑HAR~\cite{reyes2016transition}, MotionSense~\cite{malekzadeh2019mobile}, WISDM~\cite{weiss2019smartphone}, KU‑HAR~\cite{sikder2021ku}, and RealWorld~\cite{sztyler2016body}—to enable controlled cross-dataset evaluation of OOD methods.
Nevertheless, the HAR domain still lacks a unified benchmark comparable to DomainBed~\cite{gulrajanisearch} in computer vision. 
Many studies report results only in proprietary settings or limited dataset combinations~\cite{lu2022local,qian2021latent,dai2024contrastsense}, and some do not release code or data—making it difficult to assess whether comparisons are truly fair or methods are reproducible across diverse users, devices, and environments~\cite{cheng2025generalized,fan2025multi,liu2025unraveling,xiong2025deconfounding,zhang2024diverse,hong2024crosshar}.
A key bottleneck is the absence of a unified, realistic benchmark to standardize evaluation across a diverse set of OOD scenarios in sensor-based HAR, resulting in fragmented and inconsistent comparisons.

Beyond the absence of a unified benchmark, a second core challenge is this: Do existing OOD methods truly deliver practical benefits in HAR, and under which specific domain-shift scenarios?
Some OOD methods—such as MLDG~\cite{li2018learning}, DANN~\cite{ganin2016domain}, and CORAL~\cite{sun2016deep}—may not actually outperform a simple ERM baseline when applied to human activity recognition (HAR)~\cite{lu2022semantic}.
Meanwhile, methods explicitly designed for HAR, such as AFFAR~\cite{qin2022domain}, have shown modest gains by blending domain‑invariant and domain‑specific representations—but still fall short of consistently improving across all domain shifts.
This raises pressing questions: Which OOD techniques work reliably for which HAR scenarios? 
Researchers need a comprehensive study that evaluates a spectrum of OOD-HAR scenarios and provides actionable recommendations, enabling practitioners to confidently select OOD methods tailored to specific domain-shift characteristics.


To address the above challenges, we propose \benchname, an open-source benchmark and testbed for out-of-distribution human activity recognition. 
\benchname includes six publicly available time-series sensor datasets and supports four domain-shift scenarios: cross-person, cross-position, cross-device, and cross-time. 
It implements sixteen comparative methods, each using both CNN-based and Transformer-based architectures, and incorporates two model selection protocols. 
Our extensive experiments reveal the following findings:
No single method dominates across all scenarios, emphasizing the need for hybrid models and nuanced architecture design;
When unsure what method to choose, CORAL, Fish, and Fishr may be good options;
Model architecture choice (CNN vs. Transformer) should be tailored to task characteristics for optimal OOD performance;
Adaptive model-selection mechanisms, such as meta-learning or importance weighting, show promise in improving performance under domain shifts;
The accuracy of different methods varies across classes, suggesting that a combination of methods could lead to improved overall performance, particularly when considering misclassification patterns and class-specific performance.
We release \benchname as an open platform for reproducible evaluation and flexible extension. By launching it as an open platform, we aim to foster fair, reproducible evaluations and accelerate the progress of real-world activity-recognition systems.

Our contributions are summarized as follows:
(1) We propose \benchname, the first unified benchmark for OOD generalization in HAR.
(2) We define four realistic domain shift settings: cross-person, cross-position, cross-dataset, and cross-time.
(3) We comprehensively evaluate 16 OOD algorithms, including both CV-derived and HAR-specific methods, under both CNN and Transformer backbones. While we focus on CNN and Transformer architectures, our benchmark is open to any base model.
(4) We provide a detailed analysis of algorithm selection strategies, class-level behavior, and computational costs. Additionally, we offer insightful findings, practical recommendations, and discuss future research directions to guide the development of HAR systems.

\section{Related Work}
\label{sec-main-related}

\subsection{Human Activity Recognition}

Machine learning for sensor-based human activity recognition (HAR) is essential for applications such as fall detection in healthcare, assisted living, and interactive fitness systems~\cite{demrozi2020human,gu2021survey}. 
Early HAR methods primarily employed traditional algorithms like SVM, k‑nearest neighbors (k‑NN), decision trees, hidden Markov models (HMMs), and conditional random fields (CRFs), relying on hand-engineered features from accelerometer and gyroscope signals~\cite{ramasamy2018recent}. 
Over the last decade, deep learning architectures—including convolutional neural networks (CNNs), recurrent neural networks (RNNs), LSTMs, GRUs, and CNN–RNN hybrids—have become prevalent for automatically extracting spatial–temporal features, dramatically improving accuracy on benchmark datasets~\cite{wang2019deep,fawaz2019deep,wen2023transformers,wutimesnet}. 
Despite these advances, models often experience significant performance degradation in realistic settings due to distribution shifts stemming from variations in sensor placement, user biomechanics, device hardware, and time-related sensor drift~\cite{lu2022semantic,qin2023generalizable}. 
Consequently, ensuring robustness to such distributional heterogeneity remains a major unresolved challenge in HAR research.

Recent breakthroughs in LLM‑based HAR, such as HARGPT~\cite{ji2024hargpt}, which enabled zero‑shot classification of raw IMU data via prompt based querying, and SensorLLM~\cite{li2024sensorllm}, a two‑stage sensor‑to‑text alignment framework that achieved state‑of‑the‑art HAR performance and strong cross‑dataset generalization using standard benchmarks—mark a pivotal shift in the field~\cite{haresamudram2024large,chen2024towards}. 
In this context, benchmarking small CNN and Transformer models, like ours, remains highly valuable: such lightweight models can be deployed at the edge with minimal resource use, evaluated directly on publicly available datasets, and integrated into real systems at much lower cost than switching to LLM‑based approaches~\cite{teng2023pdges,hong2024crosshar}. 
Algorithmic adaptation to LLMs is theoretically feasible, but entails significant prompt engineering and infrastructure overhead, compared to our small‑model approach with minimal adaptation cost. 
Moreover, our benchmarking framework and datasets also readily support future upgrades to larger LLM‑based models if resources allow—positioning our work as both practical today and extensible toward tomorrow’s LLM‑powered HAR.

\begin{table*}[!t]
\centering
\caption{Comparison of related paradigms}
\vspace{-.1in}
\label{tab:paradigms}
\resizebox{0.8\textwidth}{!}{
\begin{tabular}{lccccp{4cm}}
\toprule
Paradigm & Source & Target Data Pre‑test & Test‑time Adaptation & Core Goal \\
\midrule
OOD Generalization      & \cmark multiple domains & \xmark & \xmark & Learn invariant model across unseen domains. \\
Domain Adaptation         & \cmark source domains   & \cmark target (unlabeled/labeled) & \xmark & Align to the target distribution during training. \\
Source‑Free DA            & \cmark source domains   & \cmark target (unlabeled)         & \cmark before test & Adapt to target without accessing source data. \\
Test‑Time Adaptation      & \cmark source domains   & \cmark target during inference    & \cmark online & Dynamically adapt at inference time. \\
Multi‑task Learning       & \cmark multiple tasks   & \xmark                             & \xmark & Leverage correlations among related tasks. \\
\bottomrule
\end{tabular}}
\end{table*}

\subsection{OOD/DG}

OOD generalization aims to develop machine learning models that generalize well to unseen domains, addressing the distribution shift between training and testing data~\cite{zhou2022domain,wang2022generalizing}. 
\cite{wang2022generalizing} categorize OOD methods into three main groups: data manipulation, representation learning, and learning strategy. 
Data manipulation techniques, such as Mixup and domain adversarial training, augment training data to improve generalization~\cite{zhang2018mixup,huang2017arbitrary}. 
EDM~\cite{cao2024mixup} leveraged reverse Mixup to generate extrapolated domains for broader domain coverage, while DomCLP~\cite{lee2025domclp} combined domain-wise contrastive learning and prototype mixup to extract domain‑irrelevant features.
Representation learning focuses on learning domain-invariant features, with methods like domain-invariant neural networks and feature alignment~\cite{sun2016deep,li2018domain}. 
A Disentangled Prompt Representation~\cite{cheng2024disentangled} disentangled prompt subspaces to isolate invariant features while CLIPCEIL~\cite{yu2024clipceil} used channel refinement and image-text alignment to amplify those stable representations.
Learning strategy approaches, including meta-learning and ensemble learning, adapt models to new domains by leveraging prior knowledge~\cite{li2018learning,rame2022diverse}. 
Arithmetic meta‑learning~\cite{wang2025balanced} balanced gradient contributions across source domains while TIDE~\cite{agarwal2025tide} forced models to leverage local interpretable concepts under semantic variations.
These methods have been applied across various domains, including computer vision, natural language processing, and time series analysis. 

In the context of time series data, Out-of-Distribution (OOD) generalization presents unique challenges due to temporal dependencies and dynamic patterns~\cite{nie2022time,lu2024diversify}. 
Recent studies have proposed novel frameworks to address these challenges~\cite{jian2024tri,shi2024orthogonality,liu2025unraveling}. \cite{jian2024tri} introduced a Tri-Level Learning framework that incorporated sample-level, group-level, and data augmentation objectives, utilizing pre-trained Large Language Models (LLMs) to enhance generalization. 
\cite{shi2024orthogonality} proposed the Invariant Time Series Representation (ITSR) model, which learned orthogonal decompositions of time series data to capture invariant features across domains. 
\cite{liu2025unraveling} focused on extracting spatial-temporal and OOD patterns from multivariate time series data, enhancing model robustness to domain shifts. 
Although sensor-based human activity recognition relies on sequential data, many approaches don’t fully leverage temporal-network architectures~\cite{lu2022local,qin2023generalizable}. 
In this paper, we also forgo sequence-specific designs and instead explore several general-purpose neural network architectures—providing a fair, scalable benchmark for comparison.

In the domain of human activity recognition (HAR), addressing domain shifts is crucial for deploying models in real-world scenarios~\cite{lu2022local}. \cite{xiong2025deconfounding} proposed a two-branch framework with early-forking, inspired by causal inference, to disentangle causal and domain-specific factors in sensor data, improving cross-domain activity recognition. 
\cite{hong2024crosshar} introduced CrossHAR, a model that employed hierarchical self-supervised pretraining to enhance generalization across different HAR datasets. 
\cite{dai2024contrastsense} developed ContrastSense, a domain-invariant contrastive learning approach that addressed challenges such as domain shifts and class label scarcity in wearable sensing scenarios. 
Although OOD methods for sensor-based human activity recognition (HAR) have made progress, implementations are often developed independently—many with no publicly available source code. 
There is still no unified evaluation framework or benchmark library dedicated to OOD in sensor-based HAR, making systematic comparisons across methods difficult.

\subsection{OOD Benchmarks}
Benchmarks are foundational to the maturation of any research field, offering standardized datasets, unified evaluation protocols, and shared codebases that foster reproducibility, comparability, and cumulative progress. 
In OOD, several benchmarks have been particularly influential: 
DomainBed testbed~\cite{gulrajanisearch} featured seven multi-domain datasets, nine baseline algorithms, and three model‑selection criteria, and demonstrated that, when rigorously implemented, empirical risk minimization often matched or exceeded more complex OOD methods.
WILDS~\cite{koh2021wilds} provided a curated suite of 10 "real‑world" datasets spanning domains such as medical imaging, satellite data, and user‑generated content, equipped with clear train/validation/test splits and a unified API for rigorous OOD evaluation. 
WOODS~\cite{gagnonwoods} extended OOD benchmarking to sequential data by assembling eight or more time‑series datasets—ranging from sensor signals to brain recordings—and evaluating a variety of OOD generalization methods in a consistent temporal evaluation framework. 
And in the specific context of human activity recognition, a domain adaptation and generalization benchmark has been proposed that standardizes multiple smartphone‑based inertial sensor datasets (e.g. accelerometer and gyroscope data), aligning sampling rates, labels, user splits, and windowing to enable cross‑dataset evaluation of OOD methods~\cite{napoli2024benchmark}.

For sensor‑based human activity recognition (HAR), time‑series data bring distinctive requirements that a OOD benchmark must address. First, domain splits should reflect realistic distribution shifts—e.g. varying subjects, device positions, or environmental contexts—to simulate practical deployment. 
Second, the benchmark architecture should be extensible, allowing new datasets, modalities, or model architectures to be easily integrated. 
Third, it must support fair comparisons with unified evaluation pipeline. 
Finally, the framework should be scalable, enabling benchmarking across a growing set of domains, structural variants, and methods. 
A unified OOD evaluation library for sensor‑based HAR would thus provide a fair, systematic, and extensible platform for rigorous comparison of methods across diverse settings.

\section{Problem Formulation}
\label{sec-main-proform}
We consider a set of $S$ labeled source domains $\mathcal{D}^{tr} = \{\mathcal{D}^i\}_{i=1}^S$, where each domain $\mathcal{D}^i$ follows a joint distribution $P^i(\mathbf{x}, y)$ over input–output space $\mathcal{X} \times \mathcal{Y}$, with $\mathbf{x} \in \mathbb{R}^m$ and $y \in \{1, \dots, C\}$. 
$C$ denotes the number of classes.
We do not observe any samples from the (unlabeled) target domain $\mathcal{D}^T$ during training, and crucially $P^T(\mathbf{x}, y) \neq P^i(\mathbf{x}, y)$ for all $i = 1, \dots, S$, reflecting realistic domain shifts. 
Our goal is to learn a classifier $h: \mathcal{X} \to \mathcal{Y}$ using only $\mathcal{D}^{tr}$ such that the expected target error,
$\mathbb{E}_{(\mathbf{x}, y) \sim P^T} \bigl[ h(\mathbf{x}) \neq y \bigr]$
is minimized. 
This setup aligns with the standard definition of OOD generalization as optimizing model performance on \emph{unseen} target distributions using \emph{only} source data~\cite{wang2022generalizing, lu2022semantic}.

\paragraph{Relation to Other Paradigms}  
Our OOD formulation differs substantially from related paradigms.
In \emph{domain adaptation} (DA), unlabeled—or occasionally labeled—samples from the target domain are accessible during training to guide distribution alignment~\cite{alanov2023styledomain}. 
In \emph{source-free domain adaptation} (SFDA), while no source data is shared during adaptation, the method still relies on unlabeled target samples and a pretrained source model to refine performance on the target~\cite{zhang2022divide,jing2022variational}. 
\emph{Test-time adaptation} (TTA) further permits online adjustment of model parameters at inference time, using individual test inputs or small batches~\cite{zhao2023pitfalls,yang2024test}. 
Finally, \emph{multi-task learning} (MTL) shares inductive biases across different but related tasks to improve joint performance, whereas OOD tackles a single task across shifting \emph{domains}, emphasizing invariance to distributional changes~\cite{bao2022generative}. 
Unlike DA, SFDA, or TTA, OOD operates without any access to or adaptation on target data—before or during deployment. Instead, it must internally capture inter-source invariances and robust features to generalize to \emph{completely unseen} domains.

\begin{table}[ht]
\centering
\caption{Summary of dataset statistics}
\label{tab:dataset-stats}
\vspace{-0.5cm}
\resizebox{0.45\textwidth}{!}{
\begin{tabular}{lccccc}
\toprule
Dataset & Subjects & Activities & Sampling Rate& Sensors& Instances \\
\midrule
DSADS    & 8    & 19 & 25Hz  &3& 1,140,000\\
USC‑HAD  & 14   & 12 & 100Hz &2& 5,441,000 \\
UCI‑HAR  & 30   & 6  & 50Hz  &2& 1,310,000 \\
PAMAP2   & 9    & 18 & 100Hz &3& 3,850,505 \\
EMG     & 36  & 7 & 1000HZ &1& 33,903,472 \\
WESAD    & 15   & 4  & chest:700Hz; wrist:32–700Hz &8& 63,000,000  \\
\bottomrule
\end{tabular}}
\end{table}

\section{Datasets and Four OOD Scenarios}
\label{sec-main-datasets}

In this section, we introduce the six datasets used in \benchname—DSADS~\cite{barshan2014recognizing}, USC-HAD~\cite{zhang2012usc}, UCI-HAR~\cite{anguita2012human}, PAMAP2~\cite{reiss2012introducing}, EMG~\cite{lobov2018latent}, and WESAD~\cite{schmidt2018introducing}\footnote{Although WESAD is not a classic human activity recognition dataset, its large scale and use of sensor-based time-series data—combined with its focus on emotion recognition, which shares meaningful overlap with behavioral signals—makes it a valuable addition to our benchmark. Inspired by Diversify~\cite{lu2024diversify} and other multi-domain frameworks, we include WESAD to enrich our evaluation suite with diverse, real-world physiological and affective data.}—along with four constructed evaluation scenarios: Cross-Person, Cross-Position, Cross-Dataset, and Cross-Time.

\subsection{Datasets}

The six datasets used are DSADS, USC‑HAD, UCI‑HAR, PAMAP2, EMG, and WESAD. Table~\ref{tab:dataset-stats} summarizes their main statistics: number of subjects, activities, sampling rate, the number of sensor types, and approximate total instances.

\subsection{Four OOD Scenarios}
\begin{table}[ht]
\centering
\caption{Overview of scenarios covered by HAR DG works.}
\vspace{-.5cm}
\label{tab:method-scen}
\resizebox{0.45\textwidth}{!}{
\begin{tabular}{llccccc}
\toprule
\textbf{Method} & \textbf{Publication} & \textbf{Year} & \textbf{Cross‑person} & \textbf{Cross‑position} & \textbf{Cross‑dataset} & \textbf{Cross‑time} \\
\midrule
DI2SDiff & KDD~\cite{zhang2024diverse} & 2024 & $\checkmark$ &  &  &  \\
CCIL & AAAI~\cite{xiong2025generalizable} & 2025 & $\checkmark$ & $\checkmark$ & $\checkmark$ &  \\
DDLearn & KDD~\cite{qin2023generalizable} & 2023 & $\checkmark$ &  &  &  \\
LAG & ICASSP~\cite{lu2022local} & 2022 & $\checkmark$ & $\checkmark$ &  &  \\
Diversify~\cite{lu2024diversify}& TPAMI & 2024 & $\checkmark$ &$\checkmark$  & $\checkmark$ &  \\
ContrastSense & IMWUT~\cite{dai2024contrastsense} & 2024 & $\checkmark$ & $\checkmark$ & $\checkmark$ &  \\
CSI‑HAR & IEEE TMC~\cite{fan2025multi} & 2025 &  $\checkmark$&  & $\checkmark$ &  \\
CrossHAR  & IMWUT~\cite{hong2024crosshar} & 2024 &  &  & $\checkmark$ &  \\
DFDCA & TBD~\cite{cheng2025generalized} & 2025 &  $\checkmark$&  &  &  \\
STOP & CPAWC~\cite{liu2025unraveling} & 2025 &  &  &  & $\checkmark$ \\
\bottomrule
\end{tabular}}
\end{table}

We consider four real-world domain-shift scenarios: cross-person, cross-position, cross-dataset, and cross-time generalization. 
Each scenario reflects a common source of variability (differences across individuals, sensor locations, data sources, or temporal segments) that can change the data distribution. 
As shown in \tableautorefname~\ref{tab:method-scen}, these four scenarios encompass the majority of settings considered in the existing literature.
To simulate these shifts, we preprocess the raw time-series (sliding-window segmentation and normalization) and split the data into separate domains accordingly. The following description details each scenario and the dataset-specific processing steps. 
Detailed statistics can be found in \tableautorefname~\ref{tab:dataset-four-scen-stat} and illustrative diagrams can be found in \figureautorefname~\ref{fig:frame-scenari}.

\begin{table}[ht]
\centering
\caption{Statistical information for the four scenarios, including data splits, datasets, categories, input dimensions, and more.}
\vspace{-0.5cm}
\label{tab:dataset-four-scen-stat}
\resizebox{0.45\textwidth}{!}{
\begin{tabular}{cllllllllll}
\toprule
\multicolumn{1}{l}{Scenarios} & Datasets & Splits                                                           & Dimensions & Classes & Domains & Scenarios                                       & Datasets & Dimensions & Classes & Domains \\ \midrule
\multirow{6}{*}{Cross-Person} & DSADS    & {[}{[}0,1{]},{[}2,3{]},{[}4,5{]},{[}6,7{]}{]}                    & (45,1,125) & 19      & 4       & Cross-Position                                  & DSADS    & (9,1,125)  & 19      & 5       \\
                              & USC-HAD  & {[}{[}1,11,2,0{]},{[}6,3,9,5{]},{[}7,13,8,10{]},{[}4,12{]}{]}    & (6,1,200)  & 12      & 4       & Cross-Dataset                                   & D/U/H/P  & (6,1,50)   & 6       & 4       \\
                              & UCI-HAR  & {[}{[}0-5{]},{[}6-11{]},{[}12-17{]},{[}18-23{]},{[}24,29{]}{]}   & (6,1,128)  & 6       & 5       & \multicolumn{1}{c}{\multirow{3}{*}{Cross-Time}} & PAMAP2   & (27,1,200) & 12      & 4       \\
                              & PAMAP2   & {[}{[}3,2,8{]},{[}1,5{]},{[}0,7{]},{[}4,6{]}{]}                  & (27,1,200) & 12      & 4       & \multicolumn{1}{c}{}                            & EMG      & (8,1,200)  & 6       & 4       \\
                              & EMG      & {[}{[}0-8{]},{[}9,17{]},{[}18,26{]},{[}27,35{]}{]}               & (8,1,200)  & 6       & 4       & \multicolumn{1}{c}{}                            & WESAD    & (8,1,200)  & 4       & 4       \\
                              & WESAD    & {[}{[}0,1,2,3{]},{[}4,5,6,7{]},{[}8,9,10,11{]},{[}12,13,14{]}{]} & (8,1,200)  & 4       & 4       &                                                 &          &            &         &         \\ \bottomrule
\end{tabular}}
\end{table}


\begin{description}
\item[Cross-person:] This scenario aims to learn models that generalize across different individuals. 
We use six datasets collected from multiple subjects.
\item[Cross-position:] This scenario addresses variability in sensor placement. We use the DSADS dataset, which contains data from five body-worn sensor positions. 
\item[Cross-dataset:] This scenario tests generalization across different datasets. We take four datasets (DSADS (D), USC-HAD (U), PAMAP2 (P), UCI-HAR (H)) as four domains. 
\item[Cross-time:] This scenario captures temporal distribution shifts over time. We apply this to the EMG gesture data, PAMAP2, and WESAD datasets by dividing each time series into chronological segments. 
\end{description}

\subsection{Analysis}
\begin{table}[ht]
\centering
\caption{The distances between normalized data across the four scenarios.}
\label{tab:dataset-four-scen-dist}
\resizebox{0.45\textwidth}{!}{
\begin{tabular}{clllllllll}
\toprule
\multicolumn{10}{c}{Min-Max Nornmalization}                                                                                                                                   \\
\multicolumn{1}{l}{Scenarios} & Datasets & MMD     & Wasserstein1 & EMD      & Scenarios                                       & Datasets & MMD     & Wasserstein1 & EMD      \\ \midrule
\multirow{6}{*}{Cross-Person} & DSADS    & 0.28733 & 0.00385      & 8.79990  & Cross-Position                                  & DSADS    & 0.72886 & 0.03978      & 5.45760  \\
                              & USC-HAD  & 0.74128 & 0.02016      & 2.37423  & Cross-Dataset                                   & D/U/H/P  & 0.57595 & 0.02449      & 0.82824  \\
                              & UCI-HAR  & 0.24199 & 0.01299      & 2.57666  & \multicolumn{1}{c}{\multirow{3}{*}{Cross-Time}} & PAMAP2   & 0.48301 & 0.01363      & 4.12733  \\
                              & PAMAP2   & 0.43707 & 0.01045      & 7.93874  & \multicolumn{1}{c}{}                            & EMG      & 0.31570 & 0.01714      & 5.92020  \\
                              & EMG      & 0.17718 & 0.00341      & 3.28518  & \multicolumn{1}{c}{}                            & WESAD    & 0.31427 & 0.03422      & 6.63268  \\
                              & WESAD    & 0.24070 & 0.02782      & 5.41346  &                                                 &          &         &              &          \\ \midrule
\multicolumn{10}{c}{Standard Score Normalization}                                                                                                                             \\
\multicolumn{1}{l}{Scenarios} & Datasets & MMD     & Wasserstein1 & EMD      & Scenarios                                       & Datasets & MMD     & Wasserstein1 & EMD      \\ \midrule
\multirow{6}{*}{Cross-Person} & DSADS    & 0.02470 & 0.00539      & 51.38723 & Cross-Position                                  & DSADS    & 0.02714 & 0.01695      & 26.86555 \\
                              & USC-HAD  & 0.02864 & 0.02233      & 17.16211 & Cross-Dataset                                   & D/U/H/P  & 0.06415 & 0.05613      & 4.02594  \\
                              & UCI-HAR  & 0.02605 & 0.02216      & 15.26964 & \multicolumn{1}{c}{\multirow{3}{*}{Cross-Time}} & PAMAP2   & 0.02604 & 0.01077      & 21.50381 \\
                              & PAMAP2   & 0.03852 & 0.01499      & 48.81375 & \multicolumn{1}{c}{}                            & EMG      & 0.06169 & 0.02915      & 34.47249 \\
                              & EMG      & 0.00732 & 0.00571      & 31.36845 & \multicolumn{1}{c}{}                            & WESAD    & 0.04506 & 0.12925      & 29.50250 \\
                              & WESAD    & 0.01946 & 0.08236      & 27.52645 &                                                 &          &         &              &        \\ \bottomrule
\end{tabular}}
\end{table}
In this chapter, we analyze the domain-level distribution discrepancies present in the scene datasets generated by our benchmark. 
We quantify inter-domain distribution differences using Maximum Mean Discrepancy (MMD)~\cite{li2018domain} and Wasserstein distance~\cite{courty2016optimal}. 
\tableautorefname~\ref{tab:dataset-four-scen-dist} presents the average distances across sub-domains for various scene-benchmark datasets. \footnote{EMD (Earth Mover’s Distance)~\cite{rubner2000earth} measures the similarity between two probability distributions, quantifying the minimum “work” needed to transform one into the other.}
Given that there is currently no standardized pre-trained model for sequential data in human activity recognition, we compute distances directly on the normalized raw data. 
From the table, we can infer that the distances are influenced not only by task difficulty, but also by factors such as data dimensionality and the number of classes. 
This further highlights the diversity of our task scenarios.

\section{\benchname: A Benchmark for HAR in OOD Settings}
\label{sec-main-benchm}

\begin{figure}[!thbp]
    \centering
    \includegraphics[width=0.4\textwidth]{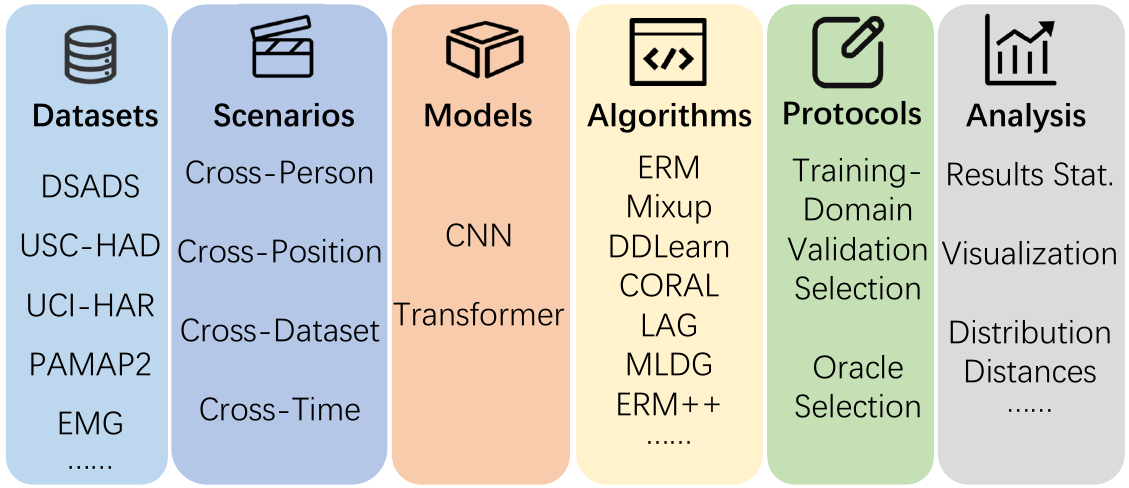}
    \caption{The components of \benchname.}
    \label{fig:frame-comp}
\end{figure}

At the heart of our large-scale evaluation lies \benchname, a PyTorch~\cite{paszke2019pytorch} testbed to streamline reproducible and rigorous research in OOD human activity recognition. 
The initial release comprises six publicly available time-series sensor datasets and supports four domain‑shift scenarios—cross‑person, cross‑position, cross‑device, and cross‑time—as well as sixteen implemented algorithms (each evaluated with CNN‑based and Transformer‑based architectures) and two model‑selection protocols. 
The components of \benchname is shown in \figureautorefname~\ref{fig:frame-comp}.
With single commands, users can run the full experimental suite. 
\benchname is a living project: we welcome contributions via pull requests, and adding a new method or dataset requires just a few lines of code.
Listing~\ref{lst:config} and \ref{lst:Usage} provide usage examples of \benchname.

\begin{lstlisting}[caption={config/test.yaml},label={lst:config}]
algorithm: 'ERM'
batch_size: 32
lr: 0.01
test_envs: [0]
output: 'output'
max_epoch: 150
task: 'cross_people'
dataset: 'dsads'
\end{lstlisting}

\begin{lstlisting}[caption={usage},label={lst:Usage}]
from core import train
# 1
filepath = './config/test.yaml'
results = train(config=filepath)
# 2
config_dict = {'algorithm': 'CORAL','batch_size': 32}
results = train(config=config_dict)
# 3
results = train(config='./config/test.yaml', lr=2e-3, max_epoch=200)
\end{lstlisting}


\subsection{Algorithms}


\begin{table}[!thbp]
\centering
\caption{Classification of Implemented Methods}
\label{table:methods}
\resizebox{0.45\textwidth}{!}{
\begin{tabular}{lcccccccc}
\toprule
Method & ERM & Mixup & DDLearn & DANN & CORAL & MMD & VREx & LAG \\
\midrule
Year & 1999 & 2020 & 2023 & 2016 & 2016 & 2018 & 2021 & 2022 \\
Publication & NY~\cite{vapnik1998statistical} & Arxiv~\cite{yan2020improve} & KDD~\cite{qin2023generalizable} & JMLR~\cite{ganin2016domain} & ECCV~\cite{sun2016deep} & CVPR~\cite{li2018domain} & ICML~\cite{krueger2021out} & ICASSP~\cite{lu2022local} \\
Data Manu. & & $\checkmark$ & $\checkmark$ & & & & & \\
Repre. Learn & & & $\checkmark$ & $\checkmark$ & $\checkmark$ & $\checkmark$ & $\checkmark$ &$\checkmark$ \\
Strategy & & & & & & & $\checkmark$ & \\
\midrule
Method & MLDG & RSC & GroupDRO & ANDMask & Fish & Fishr & URM & ERM++ \\
\midrule
Year & 2018 & 2020 & 2020 & 2021 & 2022 & 2023 & 2024 & 2025 \\
Publication & AAAI~\cite{li2018learning} & ECCV~\cite{huang2020self} & ICLR~\cite{sagawadistributionally} & ICLR~\cite{parascandolo2021learning} & ICLR~\cite{shigradient} & ICML~\cite{rame2022fishr} & TMLR~\cite{krishnamachari2024uniformly} & WACV~\cite{teterwak2025erm++} \\
Data Manu. & & & & & & & & \\
Repre. Learn & & & & & & & & \\
Strategy & $\checkmark$&$\checkmark$ &$\checkmark$ &$\checkmark$ &$\checkmark$ &$\checkmark$ &$\checkmark$ &$\checkmark$ \\
\bottomrule
\end{tabular}}
\end{table}

We include not only most of the methods implemented in DomainBed~\cite{gulrajanisearch} but also several algorithms specifically designed for OOD generalization in the activity-recognition field~\cite{lu2022local,qin2023generalizable}. 
In addition, we have organized these algorithms into categories; the detailed classification results are presented in \tableautorefname~\ref{table:methods}.

By writing each algorithm in its own method file—conforming to a unified, fixed API—and completing standardized registration and metadata entry, our benchmark easily accommodates the integration of new algorithms. 

\subsection{Model-selection protocols}

In OOD generalization, selecting an optimal model is crucial for ensuring robust performance on unseen target domains. 
Although DomainBed proposes three model-selection strategies, including training-domain validation, leave-one-domain-out cross-validation, and test-domain (oracle) validation—most HAR studies usually adopt the first approach~\cite{gulrajanisearch,lu2022local,qin2023generalizable,xiong2025generalizable,zhang2024diverse}. 
In our work, we primarily use the training-domain validation selection and additionally adapt the oracle selection.

\paragraph{Training-Domain Validation Selection}
This approach involves partitioning each training domain into training and validation subsets. 
The validation subsets are then combined to form a comprehensive validation set. 
The model is trained using the training subsets and evaluated on the aggregated validation set. 
The hyperparameters yielding the highest validation accuracy are selected.

This method assumes that the training and test distributions are similar. However, this assumption may not hold in real-world scenarios, leading to potential overfitting to the validation set. 
Consequently, the model's performance on the test domain might not align with its validation accuracy, as the validation set may not adequately represent the test distribution.

\paragraph{Oracle Selection}
In this strategy, we directly assume that the best model adapted to the test set has been selected. 
While this approach provides an accurate estimate of the model's performance on the test domain, it is not considered a valid benchmarking methodology in OOD. 
Accessing the test domain during model selection can lead to information leakage, rendering the evaluation optimistic and potentially biased.
Obviously, this is unattainable in real-world settings. Our main purpose is to evaluate the best possible performance the method can achieve within a given parameter range.

In summary, while both strategies are employed in OOD, each has its limitations. The training-domain validation selection method may not generalize well to unseen domains due to distributional differences, whereas the oracle selection approach, though more accurate, compromises the integrity of the benchmarking process. 
Researchers and practitioners should carefully consider these factors when selecting models for OOD tasks.

\subsection{Implementation Choices}

Here, we provide implementation details to help researchers better understand our benchmark and facilitate their own extensions.

\paragraph{Model Architecture}
We evaluate different algorithms using two network architectures—a CNN and a Transformer. 
For CNN, we use a shared feature-extractor composed of two sequential blocks—each block consisting of a 2D convolution layer, batch normalization, ReLU activation, and a MaxPool2d layer with kernel size $(1,2)$ and stride 2—as in \cite{lu2024diversify}, unless otherwise specified. 
For tranformer, we employ a Transformer-based feature encoder built upon an embedding layer that projects the raw input~$\mathbf{X}\in\mathbb{R}^{B\times K\times d_{\text{input}}}$ into a $d_{\text{model}}$-dimensional space, followed by a stack of encoder blocks—each block comprising multi-head self-attention (with $r$ heads using separate learned projections for Q, K, and V) and a position-wise feed‑forward sublayer, both wrapped with residual connections and layer normalization—and optionally augmented with positional encodings before attention; the final encoded sequence is flattened into a vector of size $B\times (K\cdot d_{\text{model}})$ for downstream tasks~\cite{vaswani2017attention}.
Because input dimensions vary across tasks, we personalize certain components (e.g., convolutional kernel sizes) accordingly. For some methods, we introduce minor structural modifications—for instance, LAG~\cite{lu2022local} requires an additional branch to capture global features—while keeping the core backbone unchanged to ensure a fair comparison across approaches.
By adding new network modules and making small adjustments to the common invocation mechanism, developers can seamlessly add novel network architectures, while preserving consistency in core backbone calls.

\paragraph{Hyperparameters}
We perform hyperparameter tuning using a structured grid search approach combined with a train/validation split and multiple trials to ensure robust performance estimates. Specifically, data from each task is divided with an 80\%/20\% split for training and validation. 
We fix the maximum number of epochs at 150 and use the Adam optimizer with a weight decay of $5\times10^{-4}$. 
To explore hyperparameters, we enumerate 20 combinations—for example, for ERM we vary the learning rate over $\{0.001,0.005,0.01,0.05,0.1\}$ and batch size over $\{32,64,128,256\}$. 
Each parameter setting is evaluated with three independent runs, and we report the average of the best performances across trials. 
In a four-domain cross-person scenario such as DSADS, this results in a total of $20\times3\times4 = 240$ training runs per algorithm. 
Final reported performance metrics are the mean of the best validation scores obtained over three trials.

\section{Experiments and Analysis}
\label{sec-main-exp}
We run experiments across all algorithms, datasets, and model selection criteria in \benchname. 
For each task, we consider every leave-one-domain-out configuration, training on all but one domain and testing on the held-out one, mirroring the setup from \cite{gulrajanisearch}. 
To account for random variation, we repeat the entire experimental protocol three times—reinitializing dataset splits and training seeds each run—and report the mean and standard error of task accuracy. 
For the evaluation, we not only report the average test accuracies, but also provide a ranking-based comparison to account for varying task difficulties and performance disparities across methods. 
Specifically, we rank all methods on each task and sum these ranks to compute a final score, which is then used to determine the overall ordering of methods.
This experimental protocol amounts to training a total of $16\times(3\times 60\times 4+60\times 5+60\times 4+60\times4\times5+60\times5)\times2=88320$\footnote{16 algorithms; 3×60×4 for three cross-time tasks (4 domains, 60 runs each); subsequent terms represent cross-position, cross-dataset, and cross-person tasks (DSADS, USC-HAD, PAMAP2, EMG, WESAD, HAR); the final 2 indicates two model backbones.} neural networks. 

\subsection{Findings}
\begin{figure*}[!thbp]
    \centering
    \subfigure[]{
        \includegraphics[height=0.11\textwidth]{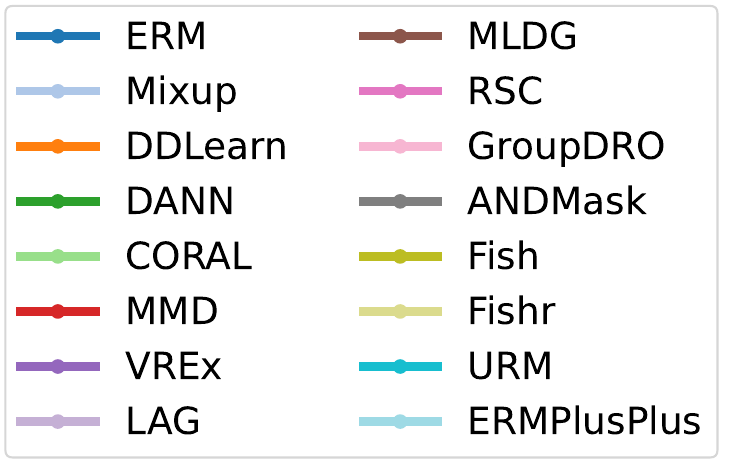}
        \label{fig:res-a}
    }
    \subfigure[]{
        \includegraphics[height=0.14\textwidth]{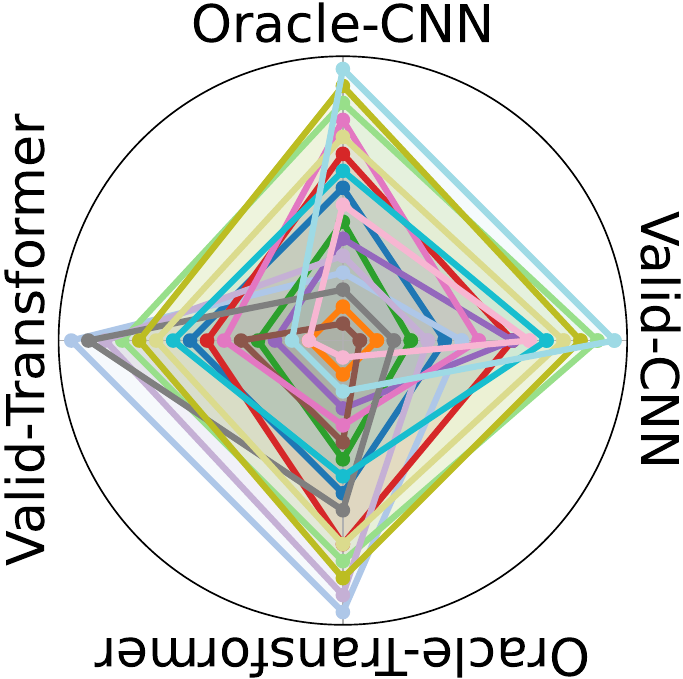}
        \label{fig:res-b}
    }
    \subfigure[]{
        \includegraphics[height=0.14\textwidth]{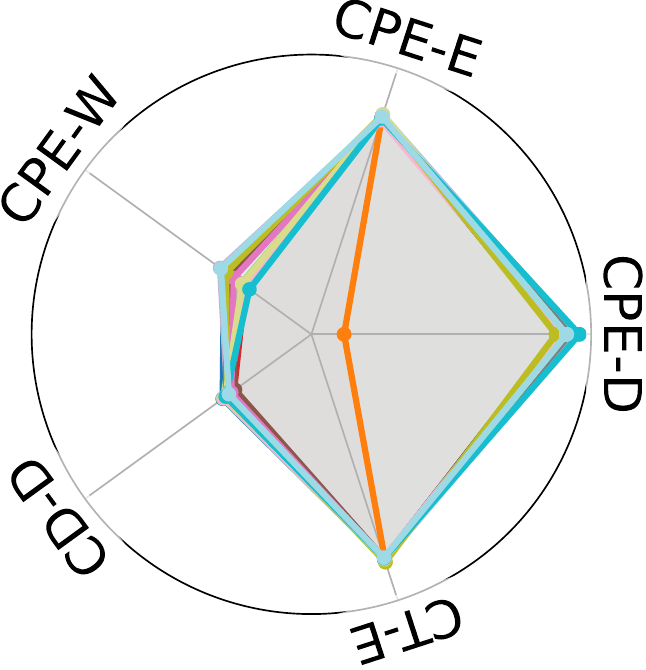}
        \label{fig:res-c}
    }
    \subfigure[]{
        \includegraphics[height=0.14\textwidth]{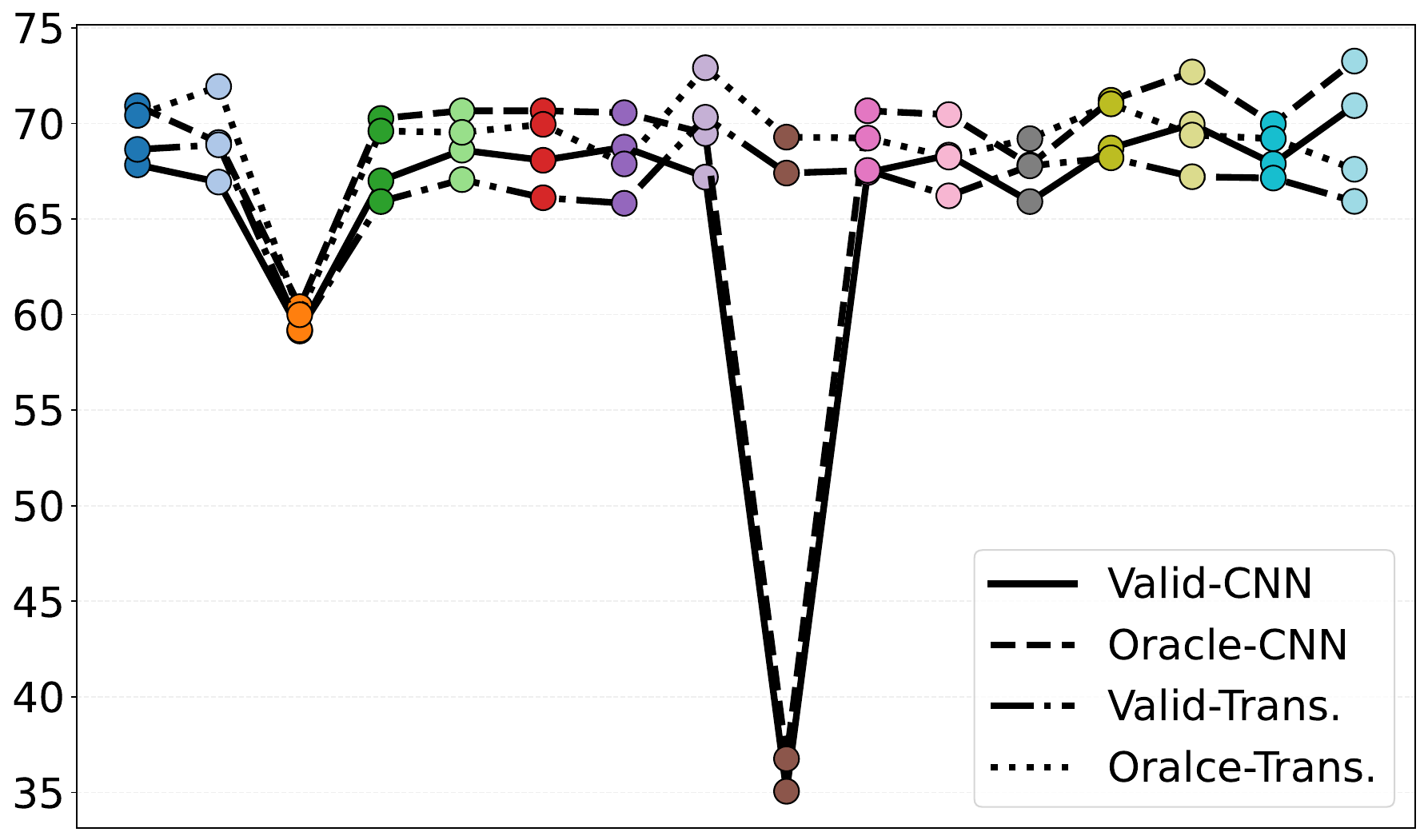}
        \label{fig:res-d}
    }
    \subfigure[]{
        \includegraphics[height=0.14\textwidth]{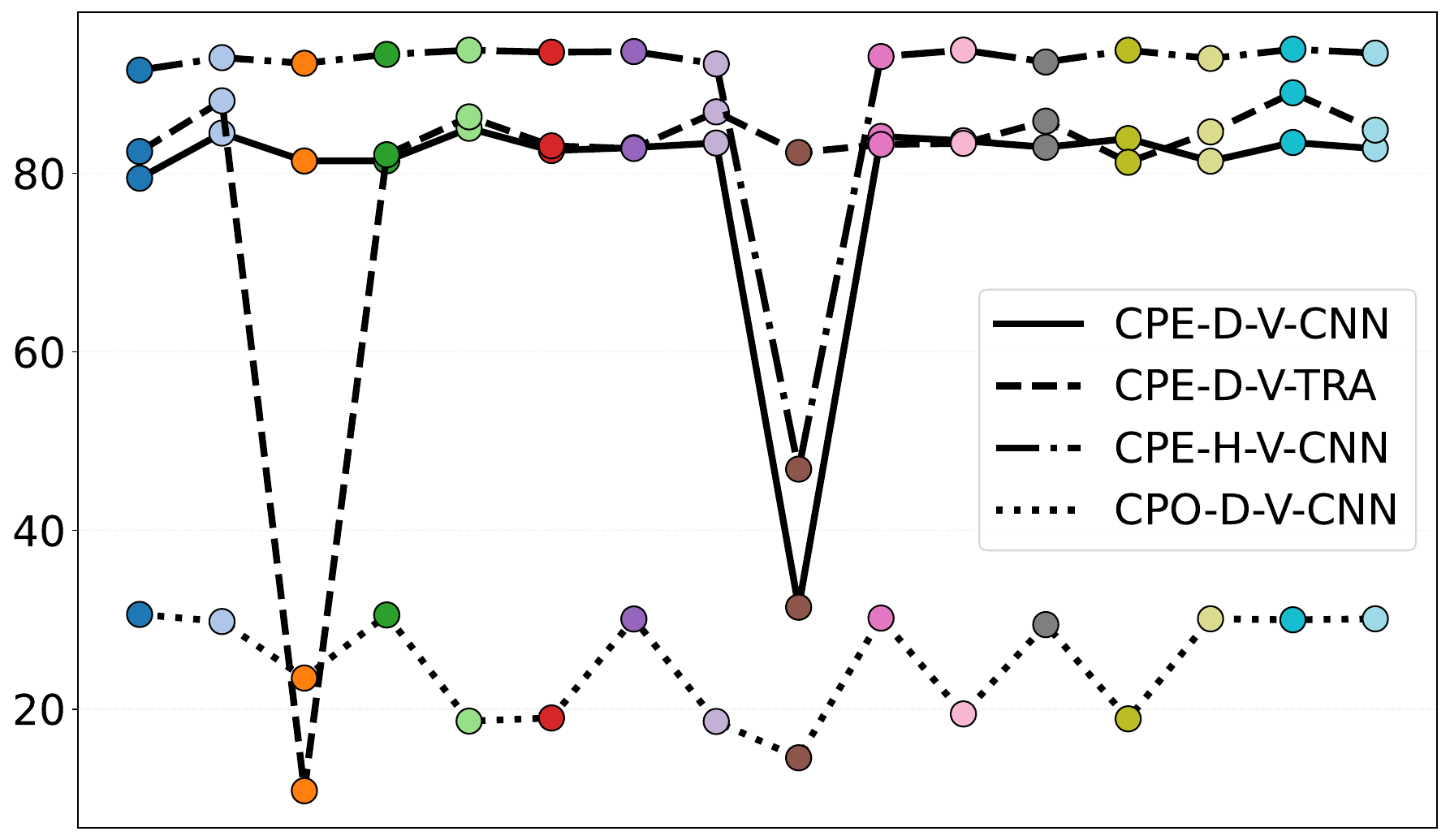}
        \label{fig:res-e}
    }
    \vspace{-0.3cm}
    \caption{This figure presents several results from our benchmark. On the far left is the legend (\figureautorefname~\ref{fig:res-a}), indicating the color corresponding to each algorithm.
\figureautorefname~\ref{fig:res-b} shows the overall ranking of different algorithms across various selection strategies and model architectures.
"Valid-CNN" refers to the results obtained using Training-Domain Validation Selection with a CNN architecture. 
\figureautorefname~\ref{fig:res-c} illustrates the performance of each algorithm across different scenarios; for instance, "CPE-W" refers to the WESAD dataset under the cross-person scenario. All results here are based on the CNN architecture with Training-Domain Validation Selection.
\figureautorefname~\ref{fig:res-d} presents the average performance of each algorithm across tasks, considering different model architectures and selection strategies.
\figureautorefname~\ref{fig:res-e} displays the performance of each algorithm under specific settings; for example, "CPE-D-V-CNN" represents the performance on the DSADS dataset in the cross-person scenario using CNN with Training-Domain Validation Selection. The other notations follow the same format.}
    \label{fig:res}
    \vspace{-0.5cm}
\end{figure*}

\figureautorefname~\ref{fig:res} summarizes the results of \benchname. 
For each task, we evaluated multiple algorithms, model-selection criteria, and model architectures. 
Specifically, we report the average out‑of‑distribution test accuracies and ranks over all combinations of dataset, algorithm, and model-selection criterion included in the initial release of \benchname. 
These experiments encompass sixteen widely used OOD algorithms under identical conditions.
The evaluational process is repeated over three independent runs using the same fixed hyperparameter set, and results are reported as mean ± standard error.
From results, we draw some major conclusions:

\paragraph{No method excels on all tasks} As shown in \figureautorefname~\ref{fig:res-c} and \ref{fig:res-e}, CORAL achieves the best performance on the cross-person DSADS task; URM performs best on UCI-HAR; and ERM achieves the highest accuracy in cross-position tasks. Although ERM++ dominates on cross-time generalization using CNN as the backbone, its performance becomes unsatisfactory when the backbone is switched in \figureautorefname~\ref{fig:res-b}. 
For consistent results, methods like CORAL, Fish, and Fishr may be preferable. 
This suggests that selecting different algorithms or backbones depending on task characteristics could be a promising direction for future research.

\paragraph{The choice of backbone model significantly influences results} In \figureautorefname~\ref{fig:res-b}, under a CNN backbone, ERM++ secures the top performance, while ANDMask and LAG rank 14th and 12th, respectively. 
However, when switching to a Transformer backbone, ERM++ drops to 14th, whereas ANDMask and LAG rise to the top three positions. 
For MLDG, the impact of the backbone is even more pronounced—likely because the limited grid search under CNN did not locate optimal hyperparameters.

\paragraph{Oracle-based selection outperforms training-domain validation}
Unlike DomainBed, we do not split off part of the test set for model selection (since that would compromise OOD validity). 
Instead, we assume access to optimal parameters (“Oracle”) to evaluate maximal model potential. 
As shown in \figureautorefname~\ref{fig:res-b} and \ref{fig:res-d}, Oracle selection consistently outperforms training-domain validation by nearly 2 percentage points. 
This highlights the importance of intelligent hyperparameter tuning. Interestingly, the ranking order across methods remains largely consistent regardless of selection strategy, although some local reorderings occur—for example, ERM++ is top-ranked in both Valid-CNN and Oracle-CNN, while ERM moves from 11th to 8th.

\paragraph{OOD methods designed for activity recognition do not necessarily maintain stability}
Across all tasks, CNN-based LAG and DDLearn rank 12th and 15th; Transformer-based LAG and DDLearn rank 3rd and 15th (as shown in \figureautorefname~\ref{fig:res-b}. 
These results fall short of the dominance reported in their original papers. 
This discrepancy may stem from the modifications we made for consistency and fairness in our unified codebase. 
We welcome authors to contribute their own implementations to enhance our benchmark.

\paragraph{Different scenarios and model architectures lead to significant variability in algorithm performance}
As shown in \figureautorefname~\ref{fig:res-c} and \ref{fig:res-e}, the best Transformer-based method on DSADS exceeds the best CNN-based method by 4 points; conversely, on WESAD (cross-time), the top CNN-based approach outperforms the best Transformer. 
This illustrates the necessity of matching network architecture to task. 
On UCI-HAR, algorithmic performance converges with generally high accuracy across the board. 
However, on cross-position tasks, performance disparities are substantial—demonstrating that task difficulty must be taken into account and that simple averaging of accuracies across different tasks may be misleading.

\begin{figure}[htbp]
    \vspace{-.6cm}
    \centering
    \subfigure[ERM]{
    \includegraphics[width=0.22\textwidth]{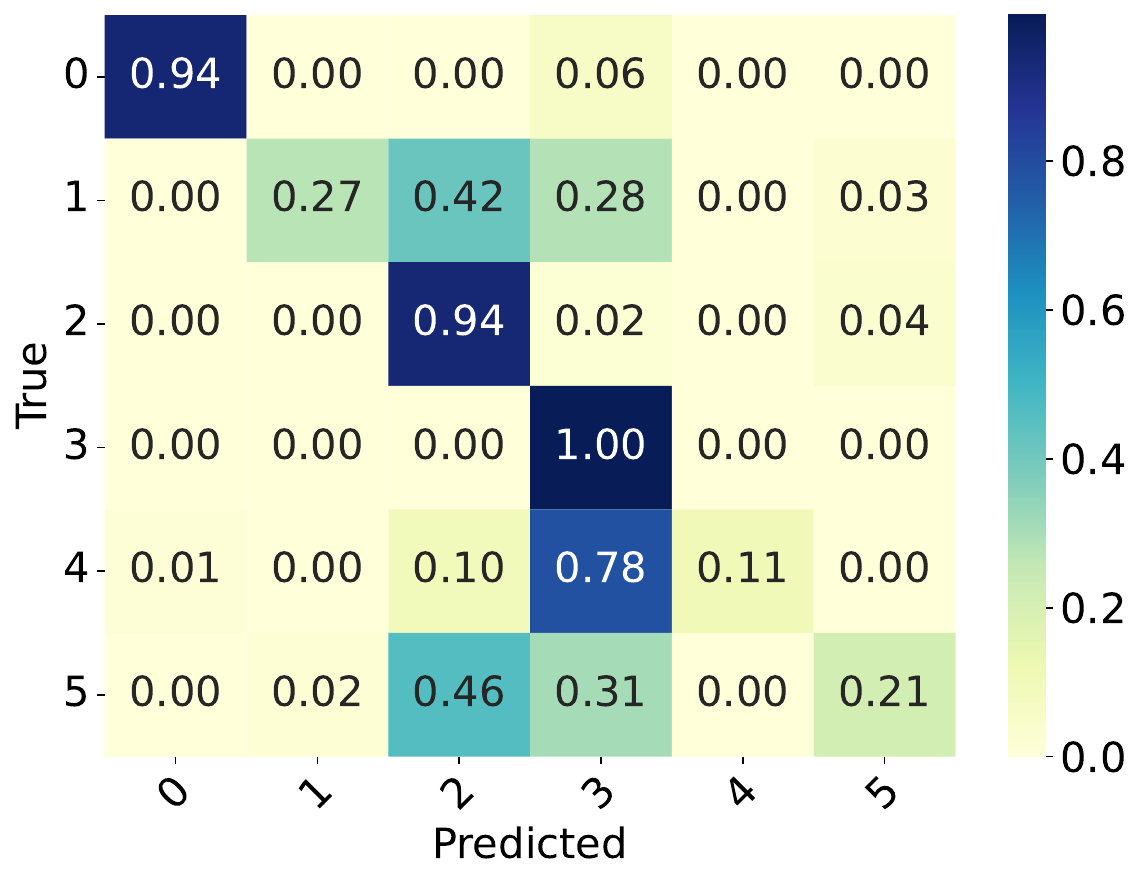}
    \label{fig:emg-conf-erm}
    }
    \subfigure[LAG]{
    \includegraphics[width=0.22\textwidth]{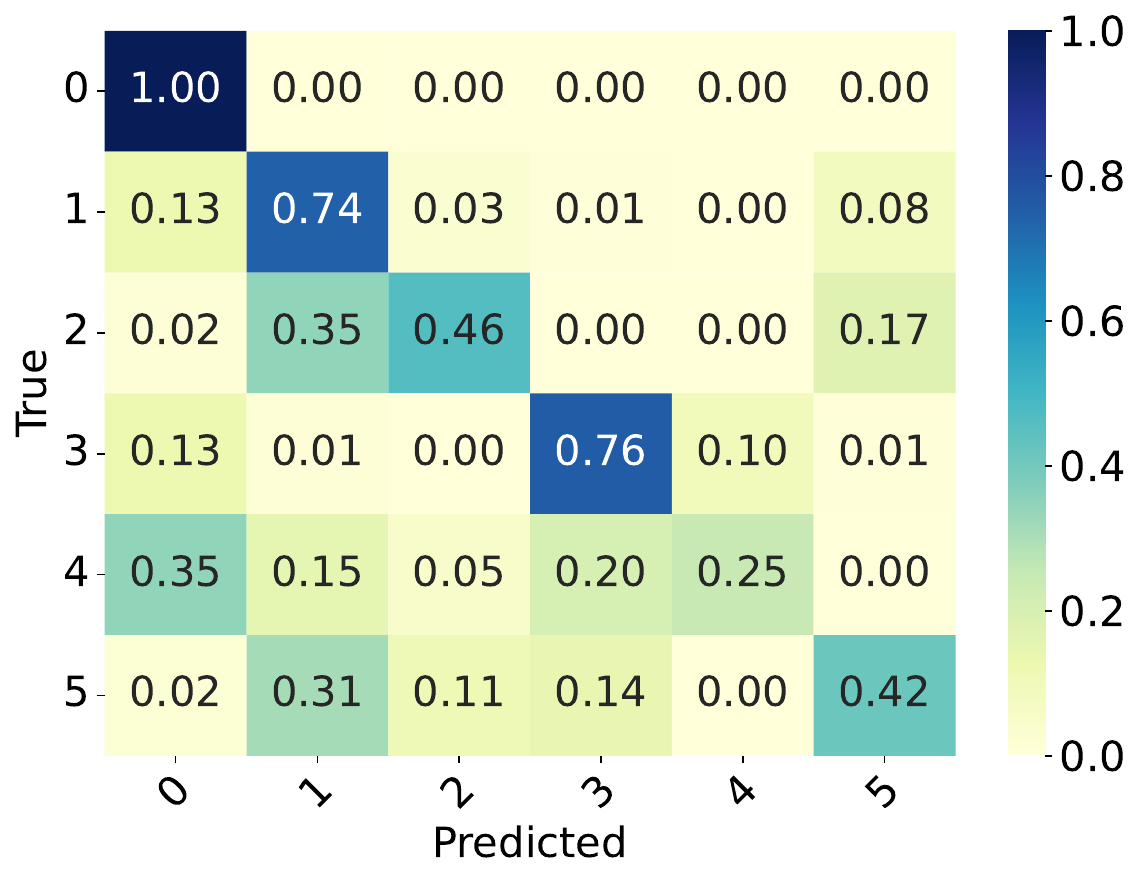}
    \label{fig:emg-conf-lag}
    }
    \vspace{-.5cm}
    \caption{Confusion matrices using CNN with validation selection on the first task of EMG.}
    \label{fig:emg-confusion}
    \vspace{-.5cm}
\end{figure}

\paragraph{The accuracy of different methods varies across classes}
As shown in \figureautorefname~\ref{fig:emg-confusion}, ERM and LAG focus on different class patterns—ERM performs well on classes 2 and 3, whereas LAG excels on classes 0, 1, and 5. 
Moreover, their misclassification patterns differ: classes are often misclassified into those that are better recognized by the model. 
This observation suggests that our benchmark should incorporate more diverse evaluation metrics in the future. 
It also implies that combining different methods could potentially lead to improved overall performance.

\begin{figure}[!thbp]
    \centering
    \includegraphics[width=0.48\textwidth]{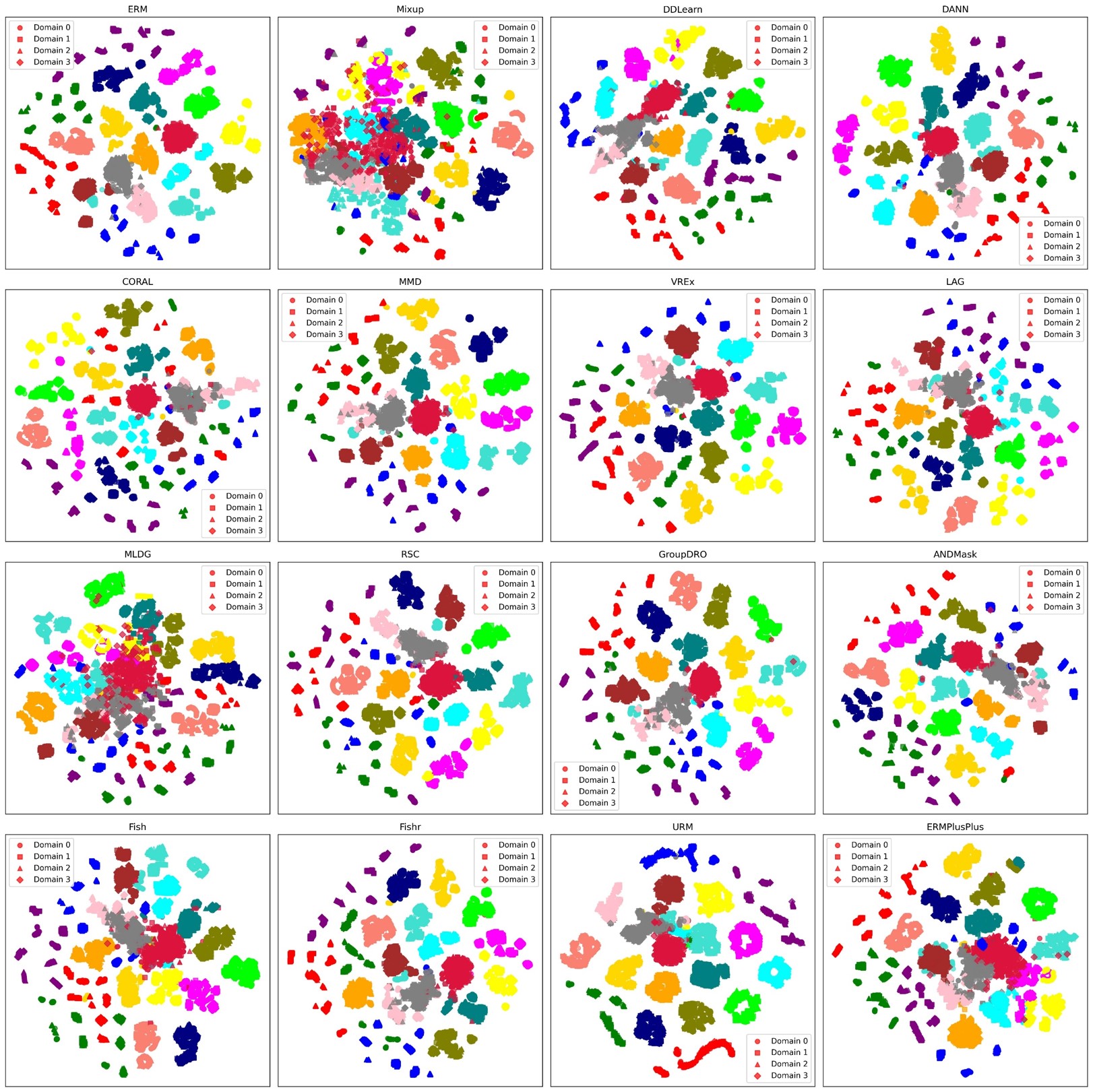}
    \vspace{-0.6cm}
    \caption{t-SNE visualization of the DSADS dataset under the cross-person setting using a CNN-based model.}
    \label{fig:vis-cnn-cpe}
    \vspace{-0.6cm}
\end{figure}

\paragraph{Sensitivity to window length varies across methods}
\begin{table*}[!htbp]
\centering
\caption{Results under the cross-time setting with different window lengths (EMG).}
\label{table:app-windleg}
\vspace{-0.4cm}
\resizebox{\textwidth}{!}{
\begin{tabular}{lllllllllllllllll}
\toprule
Window Length & ERM   & Mixup & DDLearn & DANN  & CORAL & MMD   & VREx  & LAG   & MLDG  & RSC   & GroupDRO & ANDMask & Fish  & Fishr & URM   & ERM++ \\ \midrule
100                 & 75.37 & 73.93 & 51.16   & 73.36 & 76.98 & 76.56 & 62.65 & 75.91 & 25.41 & 76.55 & 76.79    & 75.71   & 77.73 & 77.43 & 63.78 & 64.71 \\
200                 & 72.78 & 68.93 & 51.85   & 72.95 & 74.68 & 74.10 & 65.21 & 72.95 & 24.10 & 72.86 & 74.43    & 71.96   & 79.01 & 76.34 & 61.36 & 69.41 \\
500                 & 75.61 & 66.30 & 51.11   & 73.86 & 75.97 & 75.27 & 76.11 & 75.39 & 25.80 & 77.71 & 74.93    & 72.86   & 78.52 & 79.10 & 68.07 & 73.30 \\ \bottomrule
\end{tabular}}
\end{table*}

The original window was fixed for clarity, but the benchmark exposes window length as a single parameter so all methods can be re-run with arbitrary lengths. 
Here, we added experiments across multiple window sizes to demonstrate how different methods respond.
As shown in \tableautorefname~\ref{table:app-windleg}, part of the results are reported and show varying sensitivity to window length.

\paragraph{Large-model baselines do not yet outperform compact HAR models}

\begin{table}[htbp]
\vspace{-0.2cm}
\caption{The comparison results between large-model methods and small-model methods on USC-HAD.}
\vspace{-0.2cm}
\label{tab:app-llm}
\begin{tabular}{lll}
\toprule
Target & 0     & 2     \\ \midrule
HARGPT & 5.48  & 5.54  \\
ERM    & 67.08 & 64.79 \\ \bottomrule
\end{tabular}
\vspace{-.1cm}
\end{table}

We focused primarily on classic deep-learning baselines because many on-device scenarios still rely on compact models. 
In addition, small models can serve as encoders for larger systems. 
We added a large-model testing interface and initial LLM-style experiments (HARGPT~\cite{ji2024hargpt} on GPT-OSS:120b~\cite{openai2025gptoss120bgptoss20bmodel}).
As shown in \tableautorefname~\ref{tab:app-llm}, the performance of large-model methods is much lower than that of small-model methods. 

\paragraph{Model capacity and architecture jointly affect generalizatio}

\begin{table*}[!htbp]
\centering
\caption{Results with different model architectures (EMG).}
\label{table:app-arct}
\vspace{-0.4cm}
\resizebox{\textwidth}{!}{
\begin{tabular}{lllllllllllllllll}
\toprule
Architecture      & ERM   & Mixup & DDLearn & DANN  & CORAL & MMD   & VREx  & MLDG  & RSC   & GroupDRO & ANDMask & Fish  & Fishr & URM   & ERM++ \\ \midrule
Small & 66.19 & 54.52 & 45.28   & 64.1  & 70.51 & 68.2  & 71.05 & 28.06 & 76.74 & 68.09    & 64.61   & 77.51 & 74.91 & 50.96 & 62.82 \\
Mid   & 70.17 & 64.19 & 51.26   & 69.84 & 72.82 & 72.41 & 72.57 & 21.23 & 73.98 & 73.08    & 70.01   & 76.9  & 76.73 & 57.69 & 69.25 \\
Large & 71.68 & 68.39 & 69.61   & 69.55 & 72.62 & 72.52 & 61.63 & 28.86 & 73.64 & 72.19    & 72.8    & 75.98 & 76    & 68.69 & 67.93 \\
RNN   & 26.28 & 24.31 & 18.27   & 22.24 & 26.06 & 19.94 & 19.26 & 18.73 & 17.88 & 26.05    & 27.16   & 28.5  & 23.28 & 18.64 & 20.11 \\
LSTM  & 19.1  & 19.65 & 17.84   & 19.02 & 18.25 & 19.63 & 18.89 & 17.99 & 17.44 & 18.34    & 29.42   & 17.85 & 19.68 & 17.44 & 17.47 \\ \bottomrule
\end{tabular}}
\end{table*}

We also add experiments to analyze the effect of different architectures and the results are shown in \tableautorefname~\ref{table:app-arct}.
While the benchmark is primarily empirical, we now add targeted analysis: larger-capacity models benefit more from aggressive data augmentation (e.g., Mixup), whereas classic alignment methods (e.g., CORAL) remain robust for smaller models and low-data regimes. 
We hypothesize a connection between model capacity, available training data, and algorithmic generalization; to support this claim we expanded experiments across multiple model sizes.
While RNN/LSTM models can handle sequences, recent works (e.g., LAG, DDLearn) demonstrate that CNNs and Transformers achieve stronger generalization in HAR. Nevertheless, our benchmark is modular, making it easy to integrate RNNs or LSTMs.

\paragraph{Inference efficiency is comparable across benchmark methods}
\begin{table*}[!htbp]
\centering
\caption{Time consumption of 100 inferences of different methods (EMG).}
\label{table:app-timecompl}
\vspace{-0.4cm}
\resizebox{\textwidth}{!}{
\begin{tabular}{lllllllllllllllll}
\toprule
Alg     & ERM   & Mixup & DDLearn & DANN  & CORAL & MMD   & VREx  & LAG   & MLDG  & RSC   & GroupDRO & ANDMask & Fish  & Fishr & URM  & ERM++ \\ \midrule
Time(s) & 32.21 & 30.32 & 30.61   & 30.05 & 29.91 & 29.93 & 31.07 & 31.49 & 32.22 & 31.41 & 31.58    & 30.99   & 29.33 & 30.91 & 30.5 & 31.65 \\ \bottomrule
\end{tabular}}
\end{table*}
We conducted 100 inference runs on Domain~0 of the EMG dataset to measure the inference time, and the results are reported in \tableautorefname~\ref{table:app-timecompl}. 
Overall, the inference time across different benchmark methods remains at a comparable level, and no substantial gaps are observed among them, indicating similar computational efficiency.

\paragraph{F1-score largely agrees with accuracy-based comparisons}
\begin{table*}[!htbp]
\centering
\caption{Results of f1 scores from different methods.}
\label{table:app-f1score}
\vspace{-0.4cm}
\resizebox{\textwidth}{!}{
\begin{tabular}{lllllllllllllllll}
\toprule
Dataset & ERM   & Mixup & DDLearn & DANN  & CORAL & MMD   & VREx  & LAG   & RSC   & MLDG  & GroupDRO & ANDMask & Fish  & Fishr & URM   & ERM++ \\ \midrule
DSADS   & 80.06 & 80.63 & 77.92   & 82.89 & 84.32 & 83.81 & 81.59 & 84.02 & 83.17 & 31.92 & 83.47    & 83.13   & 82.45 & 80.93 & 82.79 & 82.75 \\ \bottomrule
\end{tabular}}
\end{table*}
Following DomainBed, we initially adopted accuracy as the primary evaluation metric for consistency with prior studies. 
To provide a more comprehensive evaluation, we further report the F1-score in \tableautorefname~\ref{table:app-f1score}. 
The results show that the relative performance trends measured by F1-score are largely consistent with those observed using accuracy, suggesting that the comparisons among benchmark methods are stable across different evaluation metrics.

\subsection{Visualization Study}

To better evaluate how well different methods extract features across tasks, we performed t-SNE visualizations on feature representations for both training and testing data, making comparisons across various algorithms and tasks.
\figureautorefname~\ref{fig:vis-cnn-cpe} shows the visualizations of various algorithms. 
Generally, better-performing algorithms tend to produce clearer visualizations, while poorer performance often corresponds to less distinguishable patterns.

\begin{figure}[htbp]
    \centering
    \vspace{-.3cm}
    \includegraphics[width=0.5\textwidth]{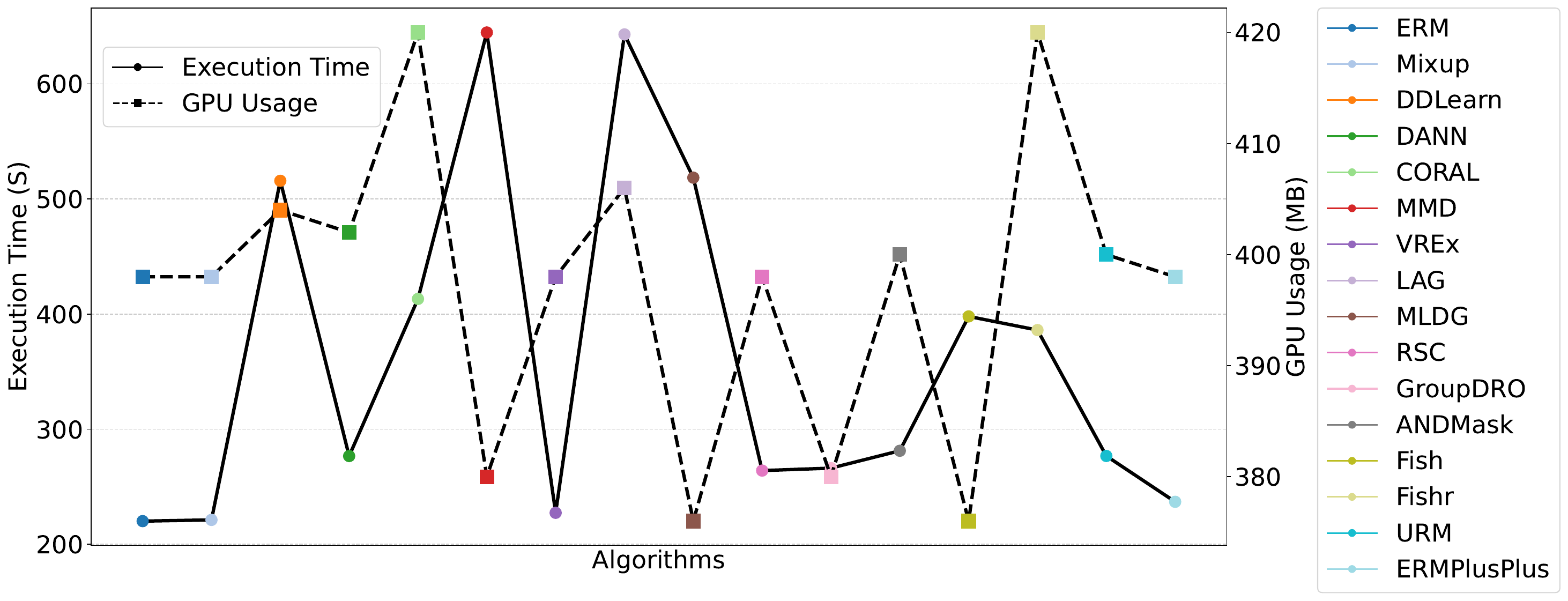}
    \vspace{-.6cm}
    \caption{Time consumption and GPU usage statistics.}
    \label{fig:resource-stat}
    \vspace{-0.6cm}
\end{figure}

\subsection{Resource Consumption Statistics}
\figureautorefname~\ref{fig:resource-stat} presents the time consumption and GPU usage (RTX 3090) of each algorithm when running a single experiment on the DSADS dataset under the cross-person setting with a CNN-based architecture. 
It can be observed that time-series data processed by simple networks runs quite efficiently. 
Among the methods, ERM achieves the highest efficiency as it involves no additional operations. 
In contrast, LAG and MMD incur longer runtimes, while DDLearn has the longest overall runtime due to its extensive preprocessing (not included in the figure). 
GPU usage is generally similar across algorithms; surprisingly, MLDG and Fish consume even less GPU memory than ERM.

\subsection{Discussion}
While no single strategy dominates across all human activity recognition scenarios, our benchmark shows that intelligently matching model architectures (CNN vs Transformer) with task characteristics, combined with dynamic algorithm selection, tends to yield the most reliable OOD performance. 
Moreover, oracle-like hyperparameter tuning—though idealized—outperforms standard validation by $\sim2\%$, underscoring the need for adaptive model-selection mechanisms (e.g., meta-learning, importance weighting) that better approximate true OOD behavior. 
Practical challenges remain in achieving robustness and reproducibility, particularly as many HAR-specific OOD methods fail to replicate their reported gains when standardized in unified codebases. 

At this point, we can attempt to address the questions raised in the Introduction:
\begin{itemize}
    \item Not all methods are effective in the context of human activity recognition (HAR); for example, MLDG and DANN often underperform.
    \item  No single method consistently outperforms others across all scenarios. Method selection should be context-specific—for instance, ERM++ tends to work best with CNN-based models, while Mixup and LAG are more suitable for Transformer-based models. When the architecture or setting is unclear, CORAL and Fish are generally safe choices.
    \item There is still a need to further explore and refine algorithm selection strategies.
\end{itemize}

According to above discussion, future work should focus on three key directions: 
(1) integrating temporal backbones~\cite{malhotra2016lstm,ma2024utsd}, more evaluation metrics and state-of-the-art OOD techniques (e.g., FOIL~\cite{liu2024time}, Moirai~\cite{woo2024unified}) to better align with sensor-based tasks; 
(2) developing online or meta-adaptive selection strategies that operate without oracle data~\cite{lu2024towards}; 
and (3) expanding benchmarks with more realistic data shifts (aging, sensor drift, synthetic domains)\cite{huang2025timedp} and explainability tools (attention, attribution) to improve trust and deployment readiness~\cite{ni2023basisformer,kim2025deltashap,ye2025domain}.

\section{Conclusion and Future Work}
\label{sec-main-concl}
OOD generalization in human activity recognition is a nuanced challenge, and our proposed \benchname benchmark illuminates this complexity. 
By conducting rigorous experiments across six public sensor-based datasets and four realistic domain-shift scenarios, and evaluating sixteen algorithms on both CNN and Transformer architectures using two model-selection protocols, we demonstrate that no single method consistently dominates every setting. 
Instead, algorithm effectiveness is closely tied to both task characteristics and the choice of backbone, and we note that, due to the massive scale of the benchmark, hyperparameter searches were necessarily limited to 20 combinations, which may disadvantage algorithms highly sensitive to their hyperparameters. 
Furthermore, model-selection strategy matters: oracle-based tuning yields notably stronger results than traditional validation-based selection, emphasizing the need for smarter hyperparameter methods. 

Embrace \benchname as a living platform, inviting contributions of novel algorithms, datasets, and evaluation protocols. 
We also support script generation, results analysis, and other full-stack research tooling to accelerate end-to-end algorithm development and deployment.
Through iterative refinement and open collaboration, we aim to transform \benchname into a central resource for community-driven progress in building robust, generalizable, and trustworthy HAR systems for diverse deployment settings.
\begin{acks}
This work was partially supported by William \& Mary Faculty Research Award and Modal Academic Compute grant. The authors acknowledge William \& Mary Research Computing for providing computational resources and/or technical support that have contributed to the results reported in this paper. URL: https://www.wm.edu/it/rc.
\end{acks}

\bibliographystyle{ACM-Reference-Format}
\balance
\bibliography{sample-base}

\appendix

\section{More Details of \benchname}
\label{sec-details-benchmark}
\subsection{Detail Descriptions of Datasets}
\label{sec-detail-datasets}

Here we provide a detailed introduction to each dataset included in \benchname.
\begin{itemize}
  \item \textbf{DSADS (Daily and Sports Activities Dataset):}  
    8 subjects performing 19 activities. Five sensors mounted on torso, arms, and legs collected tri-axial accelerometer, gyro, and magnetometer data at 25Hz. Data were segmented into $\sim9,120$ five-second windows ($\sim1.14M$ samples).

  \item \textbf{USC‑HAD (USC Human Activity Dataset):}  
    14 subjects (balanced gender, aged 21–49) performing 12 daily activities with a hip-mounted accelerometer and gyroscope at $100Hz$, yielding $\sim5.4M$ samples.

  \item \textbf{UCI‑HAR (Human Activity Recognition using Smartphones):}  
    30 subjects performing 6 activities with smartphone accelerometer and gyroscope at $50Hz$, producing over $10,000$ labeled signal segments ($\sim1M$ readings).

  \item \textbf{PAMAP2 (Physical Activity Monitoring):}  
    9 subjects performing 18 activities using 3 IMUs (wrist, chest, ankle) at 100Hz and a heart rate monitor ($\sim9Hz$), yielding $\sim3.85M$ samples.

  \item \textbf{EMG (Surface Electromyography Activity):}  
    Multi-channel sEMG signals (typically 8 channels) recorded for various movements. Public variants contain 36 subjects with per-window temporal samples (e.g., 200 timesteps).

  \item \textbf{WESAD (Wearable Stress and Affect Detection):}  
    15 subjects underwent baseline, stress, and amusement conditions. Worn devices (chest and wrist) captured ECG, EDA, EMG, respiration, temperature, and 3-axis acceleration, totaling $\sim63M$ samples.
\end{itemize}
\subsection{Detail Descriptions of Scenarios}
\label{sec-detail-scenar}

Here we provide a detailed introduction to each scenario.

\begin{description}
\item[Cross-person:] This scenario aims to learn models that generalize across different individuals. 
We use six datasets collected from multiple subjects.
For DSADS, we adopt the provider’s original split and obtain tensors of shape $45\times1\times125$, where $45=5\times3\times3$ corresponds to five sensor positions, three sensors per position, and three axes per sensor. 
For USC‑HAD and PAMAP2, we apply sliding windows of length 200 with step size 100, producing samples of size $6\times1\times200$ and $27\times1\times200$, respectively. 
For UCI‑HAR, we also uses the provider’s split, resulting in tensors sized $6\times1\times128$. 
We then randomly partition the subjects of each dataset into 4 disjoint domains(for example, the 8 subjects in DSADS are split into four groups of 2).
For EMG, we slide windows of length 200 with 100‑step (i.e. $50\%$ overlap), normalize each sample by $\tilde{\mathbf{x}}=(\mathbf{x}-\min\mathbf{X})/(\max\mathbf{X}-\min\mathbf{X})$ over all samples $\mathbf{X}$, and obtain $8\times1\times200$ samples. 
Thirty‑six subjects are randomly divided into 4 non‑overlapping domains, each with nine participants. 
For WESAD, we use chest‑worn modalities (ECG, EDA, EMG, respiration, body temperature, and 3‑axis acceleration), apply the same EMG preprocessing to data split into 4 domains.
\item[Cross-position:] This scenario addresses variability in sensor placement. We use the DSADS dataset, which contains data from five body-worn sensor positions. Each $45\times 1\times125$ windowed sample is reshaped by splitting along the position dimension into five separate samples of shape $9\times1\times125$ (3 sensors $\times$ 3 axes per position). We treat each of the five sensor positions as a distinct domain.
\item[Cross-dataset:] This scenario tests generalization across different datasets. We take four datasets (DSADS, USC-HAD, PAMAP2, UCI-HAR) as four domains. To align modalities, we select two common sensors (same body position) from each dataset and down-sample the data so that every sample has a uniform shape of $6\times1\times50$. We restrict to six activity classes that are common to all four datasets.
\item[Cross-time:] This scenario captures temporal distribution shifts over time. We apply this to the EMG gesture data, PAMAP2, and WESAD datasets by dividing each time series into chronological segments. First we apply the same sliding-window segmentation (window size 200, step 100) to obtain a sequence of windows. Then we sort these windows by time and split them into four equal parts (first through fourth quartile of the recording). Each time-based segment is treated as a separate domain. This simulates the non-stationary nature of time series, since the underlying distribution can drift over time.
\end{description}

\end{document}